\documentclass[letterpaper, 10 pt, conference]{ieeeconf}  % Comment this line out if you need a4paper

\IEEEoverridecommandlockouts                              % This command is only needed if 
\usepackage{amsmath} % assumes amsmath package installed
\usepackage{amssymb}  % assumes amsmath package installed
\usepackage{graphicx}
\usepackage{caption}
\usepackage{booktabs}
\usepackage{float}
\usepackage{cite}
\usepackage{color}
\usepackage[table]{xcolor}
\usepackage{xcolor}
\usepackage{multirow}
\usepackage{fontawesome5}
\usepackage{algorithm}
\usepackage{algorithmic}
\usepackage{hyperref}
\usepackage{fancyhdr}
\usepackage{xspace}
\usepackage{pifont}

\definecolor{darkgreen}{RGB}{34,139,34}
\definecolor{codegreen}{HTML}{76B900}
\definecolor{pc}{HTML}{643EA3}
\definecolor{tj}{HTML}{12376B}
\definecolor{te}{HTML}{50701E}

\newcommand{\our}{VertiAKD\xspace}

\title{\LARGE \bf
\our: Adaptive Off-Road Kinodynamics on \\ Vertically Challenging Terrain 
}

\author{
Tong Xu$^{1}$, Chenhui Pan$^{1}$, Francesco Cancelliere$^{1,2}$, and Xuesu Xiao$^{1}$%
\thanks{$^{1}$Tong Xu, Chenhui Pan, Francesco Cancelliere and Xuesu Xiao are with the Department of Computer Science, George Mason University, USA.}%
\thanks{$^{2}$Francesco Cancelliere is also with the Department of Electrical, Electronic and Computer Engineering, University of Catania, Italy.}%
}

\begin{document}

\maketitle
\thispagestyle{empty}
\pagestyle{empty}
% \thispagestyle{withfooter}
% \pagestyle{withfooter}

%%%%%%%%%%%%%%%%%%%%%%%%%%%%%%%%%%%%%%%%%%%%%%%%%%%%%%%%%%%%%%%%%%%%%%%%%%%%%%%%
\begin{abstract}
Off-road mobility requires autonomous mobile robots to generalize across heterogeneous vehicle fleets and continuously changing terrain conditions. Existing cross-vehicle adaptation approaches generally assume flat terrain, while terrain-aware kinodynamic models often require platform-specific data collection and retraining. To this end, we propose \textit{\our}, a unified framework for transferring and adapting off-road kinodynamic knowledge across diverse vehicles on geometrically and semantically complex terrain simultaneously. \our learns a shared mobility representation that jointly encodes vehicle configurations, trajectory transitions, and local elevation and semantic terrain features. Given limited data from a novel vehicle operating on unseen terrain, \our identifies the most relevant mobility descriptors and transfers their knowledge to initialize a terrain-aware kinodynamic model via function encoders, which is then periodically refined online from streaming observations without gradient-based retraining. We evaluate \our in the Verti-Bench simulator, built on the Chrono multi-physics engine, and on five physical configurations of the Verti-4-Wheeler platform. With only one minute of new trajectory data and associated terrain features, \our reduces long-horizon prediction error by up to 34.52\% over direct mobility descriptor transfer across diverse unseen vehicle configurations and 94.43\% over competing baselines. We further demonstrate robust closed-loop trajectory tracking in both simulation and physical experiments, highlighting the effectiveness of terrain-aware cross-vehicle knowledge transfer for accurate modeling and reliable off-road navigation. 
% \our project website can be found at \textcolor{codegreen}{\href{https://anonymous.4open.science/r/VertiAKD/}{\texttt{https://anonymous.4open.science/r/VertiAKD/}}}.

\end{abstract}
%%%%%%%%%%%%%%%%%%%%%%%%%%%%%%%%%%%%%%%%%%%%%%%%%%%%%%%%%%%%%%%%%%%%%%%%%%%%%%%%
\section{Introduction}
Accurate kinodynamic modeling is fundamental to autonomous mobile robot navigation~\cite{xiao2022motion,wang2024survey,borges2022survey}. Kinodynamic models predict how robot state evolves under applied control inputs and therefore underpin model-based planning and control. In sampling-based methods such as Model Predictive Path Integral (MPPI) control~\cite{williams2018information}, candidate control sequences are evaluated through forward rollouts generated by such models, so modeling errors compound over the planning horizon and can lead to unstable, dynamically infeasible, or unsafe trajectories~\cite{lavalle2001randomized}. Therefore, reliable navigation depends on kinodynamic models that remain accurate and continue to adapt as the robot and its operating environment change.

\begin{figure}[t!]
    \centering
    \includegraphics[width=\columnwidth]{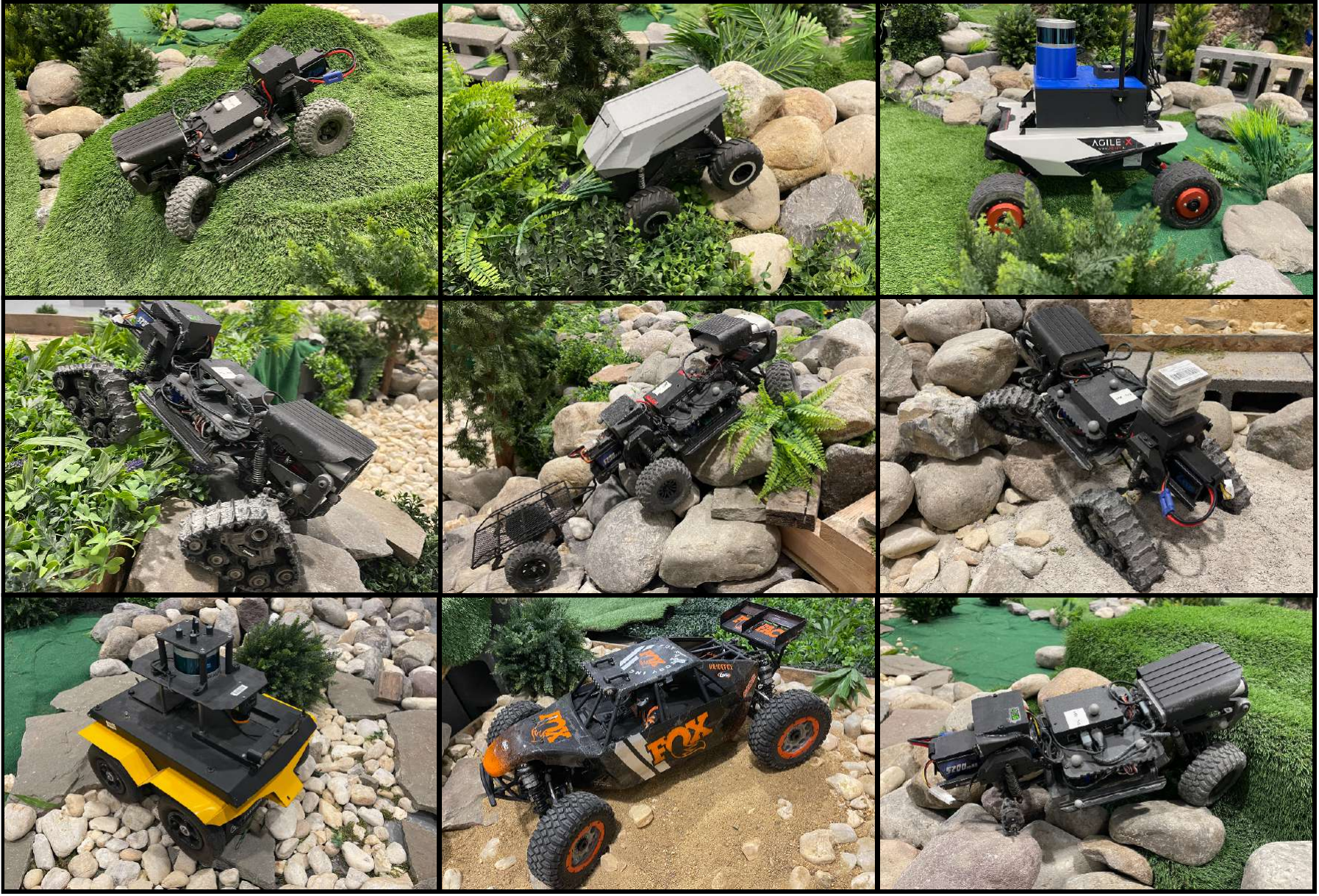}
    \caption{A heterogeneous vehicle fleet with diverse physical configurations must navigate vertically challenging terrain with changing geometry and semantics, requiring kinodynamic models that transfer across platforms and adapt periodically online.} 
    \label{fig::motivation}
    % \vspace{-15pt}
\end{figure}

To maintain prediction accuracy in unstructured off-road environments, kinodynamic models must first adapt to complex and constantly changing terrain. Vertically challenging terrain~\cite{datar2024toward} introduces abrupt variations in geometry, deformability, and surface friction, triggering pronounced roll, pitch, wheel slip, and suspension responses. Existing terrain-aware models incorporate local elevation and semantic information to capture these vehicle-terrain interactions~\cite{xiao2021learning,datar2024terrain,cai2025pietra,zhao2024physord}. Function encoders~\cite{ward2025online,ingebrand2024zero,xu2026vertiadaptor} further represent kinodynamics as a linear combination of learned neural Ordinary Differential Equation (ODE) basis functions~\cite{chen2018neural}, enabling rapid adaptation through closed-form coefficient estimation without network retraining. However, these methods are typically developed for a fixed vehicle platform, with basis functions learned for a specific set of physical parameters. Extending to a new platform therefore requires substantial platform-specific data collection and retraining, limiting the scalability of terrain-aware kinodynamic modeling across heterogeneous vehicle fleets (Fig.~\ref{fig::motivation}).

Cross-vehicle adaptation addresses fleet scalability by transferring dynamics knowledge across different platforms through shared representations or universal dynamics models~\cite{Yang2024Pushing,doshi2025scaling,xiao2025anycar}. For example, AnyCar~\cite{xiao2025anycar} learns a unified dynamics model spanning diverse vehicle embodiments, which can be rapidly adapted to a novel platform from only a few minutes of interaction data without platform-specific retraining. However, existing approaches are primarily developed and evaluated on flat or simple terrain, where dynamics variation is dominated by vehicle configuration alone. Their extension to geometrically and semantically complex terrain remains challenging, as local vehicle-terrain interaction introduces an additional, coupled source of variation.

Motivated by these limitations, we propose \textit{\our}, a unified framework for terrain-aware cross-vehicle kinodynamic transfer and online adaptation. \our\ learns a shared representation of vehicle-terrain interactions, transfers relevant mobility knowledge to novel platforms, and periodically refines the resulting kinodynamic model from streaming observations for real-time MPPI navigation. Our contributions are summarized as follows:

\begin{itemize}
    \item A shared mobility representation that jointly encodes vehicle configurations, trajectory transitions, and local terrain features to capture kinodynamic (dis)similarity across vehicle-terrain interactions;

    \item A terrain-aware function encoder that models forward kinodynamics with neural ODE basis functions incorporating elevation and semantic information, initialized by a coefficient prior constructed from relevant mobility descriptors in the shared latent space;

    \item An online adaptation scheme that refines the transferred coefficient prior from streaming observations using recursive least squares, without gradient-based retraining during deployment; and

    \item Extensive validation in the Verti-Bench simulator~\cite{xu2025verti} and on five distinct physical configurations of the Verti-4-Wheeler platform~\cite{datar2024toward}, demonstrating improved performance over state-of-the-art baselines and robust closed-loop trajectory tracking with MPPI in simulation and physical experiments.
\end{itemize}

\section{Related Work}
In this section, we review related work on terrain-aware kinodynamic modeling, meta-learning \& knowledge transfer, and online adaptation.

\subsection{Terrain-Aware Kinodynamic Modeling}
Classical off-road navigation commonly relies on simplified kinematic models, such as bicycle or ackermann formulations, or physics-based models of wheel-terrain interaction~\cite{wong2022theory}. Although these analytical models provide interpretable predictions, their parameters are difficult to identify and often do not transfer across diverse terrain surfaces and changing operating conditions.

Learning-based methods~\cite{xiao2022motion} instead estimate vehicle-terrain kinodynamics directly from data. Recent approaches incorporate geometric features, including elevation maps, surface normals, and point clouds, together with semantic or physical terrain properties to improve off-road motion prediction~\cite{xiao2021learning, atreya2022high, cai2025pietra,zhao2024physord, datar2024toward, datar2024learning, datar2024terrain}. For example, PIETRA~\cite{cai2025pietra} and PhysORD~\cite{zhao2024physord} combine learned terrain representations with physical principles to improve prediction under challenging terrain conditions. Nevertheless, these models are generally trained for a specific platform, tightly coupling its terrain representations to its kinodynamics. Applying them to a different vehicle therefore requires additional platform-specific data and model adaptation, limiting their scalability across heterogeneous fleets.

\subsection{Meta-Learning \& Knowledge Transfer}
Meta-learning aims to enable rapid adaptation from limited data. Model-Agnostic Meta-Learning (MAML)~\cite{finn2017model} learns an initialization that can be fine-tuned with a small number of gradient updates, while HyperDynamics~\cite{xian2021hyperdynamics} and RMA~\cite{kumar2021rma} infer task- or environment-specific latent representations for fast adaptation. However, gradient-based approaches require iterative backpropagation, and learned adaptation often generalize poorly to unseen vehicle-terrain interactions.

Function encoders provide an alternative by representing kinodynamics as a linear combination of learned basis functions and a compact coefficient vector~\cite{ingebrand2024zero,ward2025online, xu2026car}. This decomposition separates shared kinodynamic structure from environment-specific variation. \our\ leverages this representation to construct terrain-aware coefficient priors from relevant mobility descriptors identified in the shared latent space.

\subsection{Online Adaptation}
Online adaptation updates kinodynamic models as new observations become available. Classical methods, such as Recursive Least Squares (RLS) and Kalman filtering, provide efficient parameter updates but typically assume a fixed linear model structure. Neural approaches relax this assumption by updating network weights through online stochastic gradient descent, at the cost of repeated backpropagation that is often impractical at control frequencies. Hybrid methods, including kernel-based adaptation~\cite{ortiz2024online} and gaussian-process regression~\cite{wang2005gaussian}, balance data efficiency with generalization, but their computational and memory cost still grows with accumulated data.

\our\ performs online adaptation only in the low-dimensional coefficient space of a pretrained function encoder. During deployment, the nonlinear basis functions remain fixed, while RLS recursively refines the transferred coefficient priors from streaming observations. This enables efficient adaptation without gradient-based retraining or optimization over the accumulated dataset.

\section{Method}
\begin{figure*}[ht]
    \centering
    \includegraphics[width=2\columnwidth]{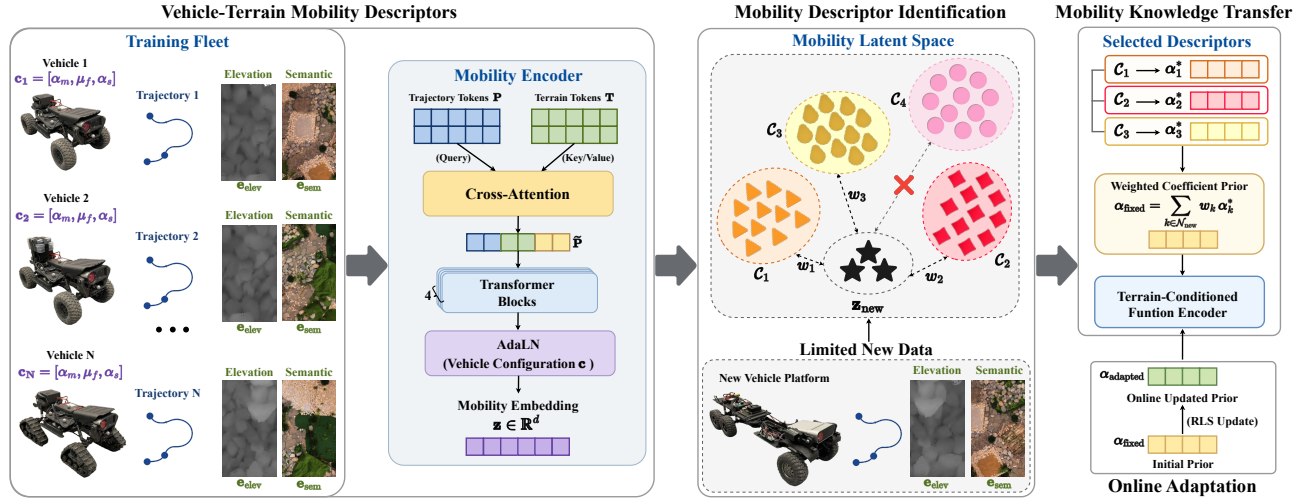}
    \caption{\textbf{\our Overview:} \textcolor{pc}{Physical configurations}, \textcolor{tj}{trajectories}, and \textcolor{te}{terrain embeddings} from a heterogeneous training fleet are jointly encoded into a shared vehicle-terrain mobility latent space using cross-attention and AdaLN (left). Limited data from a new vehicle are then used to identify relevant mobility descriptors and compute distance-based weights $w_1, w_2, w_3$ in the latent space (middle). Their coefficients are combined by the distance-based weights to initialize a terrain-conditioned function encoder (top right), whose prior is periodically refined online using RLS during deployment (bottom right).}
    \label{VertiAKD}
    % \vspace{-10pt}
\end{figure*}

Our objective is to enable rapid kinodynamic adaptation to new vehicle platforms operating on previously unseen terrain using minimal data, while periodically refining the adapted models online during deployment. We decompose \our\ into four phases: (1) learning a shared mobility latent space that jointly encodes vehicle configurations, trajectory transitions, and local terrain features (Sec.~\ref{sec:rep}); (2) identifying the most relevant mobility descriptors within this space using a distribution-aware distance metric (Sec.~\ref{sec:neighs}); (3) leveraging such prior knowledge to construct a terrain-aware coefficient prior for a terrain-conditioned function encoder (Sec.~\ref{sec:fe}); and (4) periodically refining this model online via RLS as the vehicle operates (Sec.~\ref{sec:rls}). An overview of the framework is illustrated in Fig.~\ref{VertiAKD}.

\subsection{Problem Formulation}
\label{sec:problem}

We consider a heterogeneous fleet of ground vehicles $\mathcal{V} = \{v_1, \dots, v_k\}$, each characterized by a distinct physical configuration $\mathbf{c} \in \mathbb{R}^{d_c}$, where $d_c$ denotes the number of configurable physical parameters, e.g., mass, friction, stiffness. The vehicle state at time $t$ is defined as
\begin{equation*}
    \mathbf{x}_t = \big[x_t, y_t, z_t, \phi_t, \varphi_t, \eta_t\big] \in \mathbb{SE}(3),
\end{equation*}
and possible higher-order derivatives, where $x_t, y_t, z_t$ denote the vehicle position in 3D, and $\phi_t, \varphi_t, \eta_t$ denote roll, pitch, and yaw. The control input $\mathbf{u}_t \in \mathcal{U}$ corresponds to steering and speed. Local terrain beneath the vehicle is summarized by an embedding
$
    \mathbf{e}_t = [\mathbf{e}_{\text{elev}}, \mathbf{e}_{\text{sem}}] \in \mathbb{R}^{d_e + d_s},
$
where $\mathbf{e}_{\text{elev}} \in \mathbb{R}^{d_e}$ and $\mathbf{e}_{\text{sem}} \in \mathbb{R}^{d_s}$ represent elevation and semantic features respectively. The forward kinodynamics of a single vehicle is modeled in continuous time as a neural ODE,
\begin{equation*}
    \dot{\mathbf{x}}_t = f_\theta(\mathbf{x}_t, \mathbf{u}_t, \mathbf{e}_t),
\end{equation*}
parameterized by $\theta$, which jointly captures vehicle-terrain interaction effects on the resulting motion. For compatibility with the MPPI sampling-based control framework, we adopt a discretized form,
\begin{equation}
    \Delta \mathbf{x}_t = \int_t^{t+\Delta t} f_\theta(\mathbf{x}(\tau), \mathbf{u}(\tau), \mathbf{e}(\tau)) \, d\tau,
    \label{eq:kino}
\end{equation}
with fixed time step $\Delta t$.

To enable cross-vehicle knowledge sharing, we treat $\mathcal{V}$ as the training fleet $\mathcal{V}_{\text{train}}$. For each vehicle $v_i \in \mathcal{V}_{\text{train}}$, characterized by its physical configuration $\mathbf{c}_i$, a trajectory dataset $\mathcal{D}_i = \{\tau_1, \tau_2, \dots, \tau_{M_i}\}$ is available as prior knowledge. Each trajectory
$
    \tau_j = \{(\mathbf{x}_t, \mathbf{u}_t, \mathbf{e}_t, \mathbf{x}_{t+1})\}_{t=1}^{H}
$
consists of $H$ discrete kinodynamic transition steps, each paired with the local terrain embedding observed at that step. Given a new vehicle $v_{\text{new}}$, possibly operating on terrain not represented in $\mathcal{V}_{\text{train}}$'s collected data, the problem becomes to derive the forward kinodynamics model $f_\theta$ with as little new data as possible, i.e., $\mathcal{D}_{\text{new}} = \{\tau_1, \tau_2, \dots, \tau_{M_{\text{new}}}\}$, $M_{\text{new}} \ll M_i$, with or without its physical configuration $\mathbf{c}_{\text{new}}$.

\subsection{Terrain-Aware Cross-Vehicle Mobility Representation}
\label{sec:rep}

Given trajectory transitions, vehicle configurations, and local terrain embeddings from the training fleet, we learn an encoder $\mathcal{E}_\psi$ that maps vehicle-terrain mobility interactions into a shared latent space. This structured representation preserves kinodynamic (dis)similarities across both platforms and terrain conditions, while incorporating physical priors from vehicle configurations. Such resulting structure supports downstream mobility descriptor identification and rapid adaptation to unseen vehicle-terrain interactions.

\subsubsection{Vehicle-Terrain Mobility Descriptors}
Each trajectory consists of $L$ transitions
$\mathbf{p}_t=[\mathbf{x}_t,\mathbf{u}_t,\mathbf{x}_{t+1}]$, which are tokenized and projected to dimension $d$:
$
    \mathbf{P}
    =
    [\operatorname{Proj}(\mathbf{p}_{t_0}),\dots,
    \operatorname{Proj}(\mathbf{p}_{t_0+L-1})]
    \in\mathbb{R}^{L\times d}.
$
The corresponding terrain embeddings $\mathbf{e}_t$ are tokenized in parallel to form
$\mathbf{T}\in\mathbb{R}^{L\times d}$,
aligned with the transition tokens. Within the encoder $\mathcal{E}_\psi$, the trajectory tokens $\mathbf{P}$ attend to corresponding terrain sequence $\mathbf{T}$ via cross-attention:
$
    \widetilde{\mathbf{P}} = \mathbf{P} + \operatorname{CrossAttn}\left(\mathbf{P}, \mathbf{T}, \mathbf{T}\right),
$
where trajectory tokens serve as queries and terrain tokens as keys and values. A learnable CLS token $\mathbf{h}_{\text{cls}}$ is then prepended to $\widetilde{\mathbf{P}}$ and augmented with sinusoidal positional embeddings $\mathbf{PE}$:
$
    \mathbf{H}^{(0)} = [\mathbf{h}_{\text{cls}}, \widetilde{\mathbf{P}}] + \mathbf{PE}.
$
The resulting sequence $\mathbf{H}^{(0)}$ is then processed by $N$ Transformer blocks.

Vehicle configuration is introduced through adaptive layer normalization (AdaLN)~\cite{peebles2023scalable} at the final block, allowing platform-specific attributes to refine the learned trajectory-terrain representation. The configuration vector is embedded as $\mathbf{v}_c = g_\xi(\mathbf{c})$ via a learnable encoder $g_\xi(\cdot)$ and projected into channel-wise modulation parameters via a linear layer $\Phi$:
$
    [\Delta\boldsymbol{\beta}_1, \Delta\boldsymbol{\gamma}_1, \Delta\boldsymbol{\beta}_2, \Delta\boldsymbol{\gamma}_2] = \Phi(\mathbf{v}_c),
$
which modulate the self-attention and feed-forward sublayers:
\begin{align*}
    \mathbf{H} &\leftarrow \operatorname{LN}\!\left(\mathbf{H} + \operatorname{Attn}\!\left(\mathbf{H} \odot (1+\Delta\boldsymbol{\gamma}_1) + \Delta\boldsymbol{\beta}_1\right)\right), \\
    \mathbf{H} &\leftarrow \operatorname{LN}\!\left(\mathbf{H} + \operatorname{FFN}\!\left(\mathbf{H} \odot (1+\Delta\boldsymbol{\gamma}_2) + \Delta\boldsymbol{\beta}_2\right)\right),
\end{align*}
where $\odot$ denotes element-wise multiplication, $\mathrm{LN}(\cdot)$ layer normalization, $\mathrm{Attn}(\cdot)$ multi-head self-attention, and $\mathrm{FFN}(\cdot)$ a position-wise feed-forward network. Replacing $\mathbf{v}_c$ with a learnable null embedding disables configuration modulation, yielding an unconditional pass based only on trajectory and terrain tokens. The final CLS representation produces unconditional embedding 
$
    \mathbf{z}^{u}
    =
    \mathcal{E}_\psi(\mathbf{P},\mathbf{T},\varnothing)
$
or conditional embedding
$
    \mathbf{z}^{c}
    =
    \mathcal{E}_\psi(\mathbf{P},\mathbf{T},\mathbf{c}).
$

\subsubsection{Terrain-Aware Triplet Training}
To structure the shared latent space by vehicle-terrain kinodynamic similarity, we train the encoder $\mathcal{E}_\psi$ with a terrain-aware triplet objective. For each trajectory segment $\tau$, the associated terrain embeddings are summarized as a Gaussian signature with mean $\boldsymbol{\mu}_\tau = \frac{1}{L}\sum_{t=1}^{L}\mathbf{e}_t$ and variance $\boldsymbol{\sigma}_\tau^2 = \frac{1}{L}\sum_{t=1}^{L}(\mathbf{e}_t - \boldsymbol{\mu}_\tau)^2$. Terrain dissimilarity between two segments is measured via the symmetric Kullback-Leibler (KL) divergence between their signatures, averaged across embedding dimensions. For each anchor segment $\tau_a$, we sample a positive $\tau_p$ from the same vehicle and terrain and a negative $\tau_n$ differing in vehicle configuration, terrain condition, or both, as detailed in Sec.~\ref{sec::vt_mr}. This encourages the latent space to reflect joint vehicle-terrain kinodynamic similarity, rather than vehicle identity or trajectory alone.

We apply the resulting triplets to both unconditional and conditional embeddings:
\begin{align*}
    \mathcal{L}_{u} &= \max\Big(\|\mathbf{z}^{u}_{a}-\mathbf{z}^{u}_{p}\|_2 - \|\mathbf{z}^{u}_{a}-\mathbf{z}^{u}_{n}\|_2 + \delta,\; 0 \Big), \\
    \mathcal{L}_{c} &= \max\Big(\|\mathbf{z}^{c}_{a}-\mathbf{z}^{c}_{p}\|_2 - \|\mathbf{z}^{c}_{a}-\mathbf{z}^{c}_{n}\|_2 + \delta,\; 0 \Big),
\end{align*}
where $\delta$ is a shared margin parameter. The unconditional loss $\mathcal{L}_{u}$ organizes mobility according to trajectory and terrain information alone, whereas the conditional loss $\mathcal{L}_{c}$ additionally structures the embedding by vehicle configuration. To prevent collapsed embeddings, we additionally penalize any embedding dimension whose standard deviation falls below a target $\rho$, $\mathcal{L}_{\text{var}} = \frac{1}{D}\sum_{d=1}^{D} \max\big(0,\ \rho - \sqrt{\mathrm{Var}(z_d) + \epsilon}\big)$, where $z_d$ is the $d$-th dimension of the anchor, positive, and negative embeddings. We train $\mathcal{E}_\psi$ with the combined objective $\mathcal{L} = w_u \mathcal{L}_u + w_c \mathcal{L}_c + w_v \mathcal{L}_{\text{var}}$ with $w_u$, $w_c$, and $w_v$ as constant weights.

\subsection{Distribution-Aware Descriptor Identification}
\label{sec:neighs}

After training $\mathcal{E}_\psi$, we project training fleet mobility embeddings into a low-dimensional space via Principal Component Analysis (PCA) and group them into $K$ vehicle-terrain mobility descriptors using K-means clustering, selecting $K$ by maximizing silhouette score to balance intra-cluster compactness against inter-cluster separation. This yields $\mathcal{Z}_{\mathrm{train}} = \bigcup_{k=1}^{K} \mathcal{C}_k$, where each cluster $\mathcal{C}_k$ represents a mobility descriptor shared across related vehicle-terrain interactions. Given limited data $\mathcal{D}_{\mathrm{new}}$ from a novel vehicle, we encode its trajectory segments as $\mathcal{Z}_{\mathrm{new}} = \{\mathcal{E}_\psi(\mathbf{P}_j, \mathbf{T}_j, \mathbf{c}_{\mathrm{new}})\}_{j=1}^{M_{\mathrm{new}}}$, projected into the same PCA space, using unconditional embeddings when $\mathbf{c}_{\mathrm{new}}$ is unavailable.

We compare $\mathcal{Z}_{\mathrm{new}}$ against each descriptor distribution using Sliced Wasserstein Distance (SWD)~\cite{kolouri2019generalized}, converted into normalized relevance weights, $\tilde{w}_k = \text{SWD}\left(\mathcal{Z}_{\mathrm{new}}, \mathcal{C}_k\right)^{-2}$ and $w_k = \tilde{w}_k / \sum_{j=1}^{K}\tilde{w}_j$. Descriptors are ranked by $w_k$, and the top-ranked set $\mathcal{N}_{\text{new}}$ whose cumulative weight exceeds $0.9$ is retained as the mobility descriptors used for downstream kinodynamic knowledge transfer.

\subsection{Terrain-Conditioned Function Encoder}
\label{sec:fe}

We represent forward kinodynamics (Eqn.~\eqref{eq:kino}) as a linear combination of $K_b$ neural ODE basis functions:
\begin{equation*}
    \Delta \mathbf{x}_t = \sum_{i=1}^{K_b} \alpha_i \, G_i\big(\mathbf{x}_t, \mathbf{u}_t, \mathbf{e}_t; \theta_i\big),
\end{equation*}
where each $G_i(\cdot; \theta_i)$ is computed via a fourth-order Runge-Kutta (RK4) integrator, and $\boldsymbol{\alpha} = [\alpha_1, \dots, \alpha_{K_b}]^T\in \mathbb{R}^{K_b}$ specializes the shared basis to a specific vehicle-terrain condition~\cite{xu2026vertiadaptor}. The trajectory and terrain tokens are first fused through cross-attention, and the resulting representation is provided as input to each basis function $G_i$, enabling terrain-conditioned kinodynamic modeling.

Given an example set of transitions $\{\tau_1, \tau_2, \dots, \tau_M\}$ from a single descriptor, the coefficients that best explain this data are obtained in closed form via ridge-regularized least squares,
\begin{equation} 
    \alpha^*_k = \left(G_k^T G_k + \lambda I\right)^{-1} G_k^T \Delta\mathbf{x}_k, 
    \label{eq:ls}
\end{equation}
where $G_k$ and $\Delta\mathbf{x}_k$ stack the predicted basis outputs and true state changes from the example set, and $\lambda$ and $I$ are the ridge regularization coefficient and identity matrix. We compute $\alpha^*_k$ for every descriptor identified in $\mathcal{N}_{\text{new}}$ and combine them using the weights $w_k$ from mobility latent space to obtain a coefficient prior for the novel vehicle $v_{\text{new}}$:
\begin{equation*}
    \alpha_{\text{prior}} = \sum_{k \in \mathcal{N}_{\text{new}}} w_k \, \alpha^*_k.
\end{equation*}

\subsection{Online Adaptation via RLS}
\label{sec:rls}

While $\alpha_{\text{prior}}$ provides an effective initialization, it reflects only the training fleet's prior experience and does not account for the specific conditions $v_{\text{new}}$ encounters during deployment. We therefore periodically refine the coefficient vector using RLS, warm-started from the transferred prior, as summarized in Algorithm~\ref{alg:rls}. At each control step, the observed transition $(\mathbf{x}_t,\mathbf{u}_t,\mathbf{e}_t,\mathbf{x}_{t+1})$ is appended to a fixed-size buffer $\mathcal{B}$ (lines 2-3). Once the buffer contains $B$ transitions, the basis function outputs are stacked to form $G_t$, and the corresponding observed state changes are stacked into $y_t$ (lines 4-5). The covariance is discounted by the forgetting factor $\gamma$ (line 6), after which the innovation covariance and gain are computed in closed form (lines 7-8). The coefficients are then updated using the residual $y_t-G_t\alpha_{t-1}$, followed by the covariance update (lines 9-10). The buffer is cleared for the next update cycle (line 11), while the coefficients and covariance remain fixed between updates (lines 12-13). Because the basis functions and buffer size are fixed, the cost of each update does not grow with the deployment history, enabling efficient online refinement without gradient-based retraining.

\begin{algorithm}[h]
\caption{Online Adaptation via RLS}
\label{alg:rls}
\begin{algorithmic}[1]
\STATE \textbf{Initialize:} $\alpha_0 = \alpha_{\text{prior}}$, $P_0 = \lambda^{-1}I$, $\gamma \in (0,1]$, noise $Q$, buffer $\mathcal{B} \leftarrow \emptyset$, buffer size $B$
\FOR{each control step $t = 1, 2, \dots$}
    \STATE Observe transition $(\mathbf{x}_t, \mathbf{u}_t, \mathbf{e}_t, \mathbf{x}_{t+1})$ and append to $\mathcal{B}$
    \vspace{-12pt}
    \IF{$|\mathcal{B}| = B$}
        \STATE Construct $G_t$ and $y_t$ from all transitions in $\mathcal{B}$
        \STATE $P_{t|t-B} \leftarrow \dfrac{1}{\gamma} P_{t-B}$
        \STATE $S_t \leftarrow G_t P_{t|t-B} G_t^T + Q$
        \STATE Compute gain: $K_t \leftarrow P_{t|t-B} G_t^T S_t^{-1}$
        \STATE Update coefficients: $\alpha_t \leftarrow \alpha_{t-B} + K_t(y_t - G_t\alpha_{t-B})$
        \vspace{-12pt}
        \STATE Update covariance: $P_t \leftarrow P_{t|t-B} - K_t G_t P_{t|t-B}$
        \STATE Clear buffer: $\mathcal{B} \leftarrow \emptyset$
    \ELSE
        \STATE $\alpha_t \leftarrow \alpha_{t-1}$, $P_t \leftarrow P_{t-1}$
    \ENDIF
\ENDFOR
\end{algorithmic}
\end{algorithm}

\section{Implementations}
In this section, we present implementation details of our approach and experiments.

\subsection{Vehicle Configurations and Datasets}
We evaluate \our\ in the Verti-Bench simulator~\cite{xu2025verti} and a physical testbed similar to Verti-Arena~\cite{chen2025verti} using a heterogeneous fleet operating over geometrically and semantically diverse terrain. Each vehicle configuration is represented by
$
    \mathbf{c}=[\alpha_m,\mu_f,\alpha_s]^T,
$
where $\alpha_m$ is the chassis mass scaling ratio, $\mu_f$ is the rigid tire friction coefficient, and $\alpha_s$ denotes the suspension spring stiffness scaling ratio. The training fleet contains five configurations sampled from
$\alpha_m\sim[0.5,2.0]$,
$\mu_f\sim[0.6,0.9]$, and
$\alpha_s\sim[0.6,1.8]$.

Each vehicle collects trajectories through sinusoidal random exploration over terrain with diverse elevation and semantic properties. Steering is commanded as
$\mathbf{u}_{\text{steer}}(t)=\sin(\omega_s t)$,
with $\omega_s\sim[0.1,0.5]$ Hz, while speed follows
$\mathbf{u}_{\text{speed}}(t)=v_c+A\sin(\omega_v t)$,
with $\omega_v\sim[0.1,2.5]$ Hz. The minimum and maximum speeds are sampled as $v_{\text{min}}\sim[1,2]$\,m/s and $v_{\text{max}}\sim[3,4]$\,m/s. The velocity amplitude and center are then computed as $A=(v_{\text{max}}-v_{\text{min}})/2$ and $v_c=(v_{\text{max}}+v_{\text{min}})/2$. We record current state, commanded control, next state, and vehicle-aligned $128\times128$ elevation and RGB semantic patches at $10$\,Hz, yielding terrain-conditioned transitions across diverse vehicle-terrain interactions. For terrain feature extraction, separate convolutional autoencoders with identical encoder architectures map the 2.5D elevation and RGB semantic patches to $18\times16\times16$ spatial feature maps. These features are reshaped for cross-attention in Secs.~\ref{sec:rep} and~\ref{sec:fe}. Elevation maps are globally normalized to $[-1,1]$, while RGB semantic images are normalized channel-wise to the same range. Both autoencoders are trained independently with mean-squared reconstruction loss using Adam, a learning rate of $10^{-4}$, and a batch size of $64$.

Since vehicle kinodynamics are invariant to global translation and yaw, we express the vehicle state in a gravity-aligned body frame. At each timestep, the global position and yaw are reset to zero, while roll $\phi_t$ and pitch $\varphi_t$ are retained to preserve the vehicle's orientation relative to gravity on uneven terrain. We further include the yaw rate $\dot{\eta}_t$ and longitudinal speed $v_t$ to capture rotational and momentum-dependent effects that become increasingly important at higher speeds. The resulting current state is
$
    \mathbf{x}_t
    =
    [0,\,0,\,0,\,\phi_t,\,\varphi_t,\,0,\,\dot{\eta}_t,\,v_t]^T
    \in\mathbb{R}^{8}.
$
The corresponding  next state describes the vehicle motion relative to the current body frame for vehicle position and yaw and to the gravity-aligned frame for roll and pitch:
$
    \mathbf{x}_{t+1}
    =
    [\Delta x,\,\Delta y,\,\Delta z,\,\phi_{t+1},\,\varphi_{t+1},\,\Delta\eta]^T
    \in\mathbb{R}^{6}.
$
Here, $\Delta x$, $\Delta y$, and $\Delta z$ denote the relative translation, and $\Delta\eta$ denotes the relative yaw change. In contrast, $\phi_{t+1}$ and $\varphi_{t+1}$ are represented as absolute roll and pitch angles, preserving the vehicle's orientation with respect to gravity.

\subsection{Vehicle-Terrain Mobility Representation}
\label{sec::vt_mr}

The mobility encoder consists of $4$ Transformer blocks with hidden dimension $d=32$ and $4$ attention heads. Each input trajectory window contains $L=64$ consecutive transitions $\mathbf{p}_t = [\mathbf{x}_t, \mathbf{u}_t, \mathbf{x}_{t+1}] \in \mathbb{R}^{18}$, linearly projected to $\mathbb{R}^{32}$. At each timestep, the corresponding elevation and semantic embeddings are jointly projected into a paired terrain token, forming a sequence temporally aligned with the transition tokens. This terrain sequence is fused with the transition token via cross-attention, before prepending the learnable CLS token and adding sinusoidal positional embeddings.

Vehicle configuration parameters $(\alpha_m,\mu_f,\alpha_s)$ are min--max normalized to $[0,1]$.
They are embedded into $\mathbb{R}^{32}$ by a two-layer Multi-Layer Perceptron (MLP)  with dimensions $\{3,8,32\}$ and Tanh activations. The resulting embedding modulates the final Transformer block through AdaLN with a scale factor of $0.5$. During training, the configuration embedding is replaced with a learnable null embedding
$\mathbf{e}_{\varnothing}\in\mathbb{R}^{32}$
with probability $0.1$, enabling unconditional inference when the physical configuration is unavailable.

For terrain-aware triplet sampling, each anchor trajectory window is compared with $32$ candidate windows drawn from a precomputed set for the same vehicle type. This set is periodically refreshed during training to improve trajectory diversity. Sample selection is guided by the symmetric KL divergence between the terrain signature of anchor and each candidate. Positives are drawn from the same vehicle under similar terrain (lower KL divergence), whereas negatives differ in vehicle, terrain (higher KL divergence), or both. We set the triplet margin to $\delta=2.0$ and use loss weights $w_u=1.0$, $w_c=1.0$, and $w_v=0.1$ for the unconditional, conditional, and variance-regularization terms.

The encoder is trained with Adam optimizer for up to 100K iterations using a learning rate of $10^{-4}$ and a batch size of $128$. Conditional embedding separation is evaluated on a held-out validation set every $1{,}000$ iterations, and training is terminated early if no improvement is observed for $10$ consecutive evaluations.

\subsection{Function Encoder}
\label{sec::impl_fe}

The function encoder comprises eight terrain-conditioned neural ODE basis functions. State and control inputs are independently mapped to $16$-dimensional features, while elevation and semantic embeddings are processed through separate cross-attention branches and fused into a $16$-dimensional terrain representation. The resulting features are concatenated and provided to each basis function, which uses separate MLP heads to predict 
$
    [\Delta x,\,\Delta y,\,\Delta z,\,\Delta\eta]^T
$ and 
$
    [\phi_{t+1},\,\varphi_{t+1}]^T
$
respectively.

The basis outputs are integrated using RK4 with a fixed timestep of $\Delta t=0.1$\,s. Terrain-specific coefficients are estimated via ridge-regularized least squares (Eqn.~\eqref{eq:ls}) with $\lambda=10^{-3}$. At each training step, we sample two vehicle-terrain mobility descriptors and $12$ trajectories from each descriptor. Three trajectories form the example set for coefficient estimation, while the remaining trajectories are used to optimize a $T_{\text{pred}}=16$-step rollout loss. During autoregressive rollout, terrain patches are recropped at the predicted poses and re-encoded at each step. The model is trained for $2{,}000$ gradient steps using Adam optimizer with a learning rate of $10^{-3}$.

\section{Experiments}
\label{sec:exp}

We evaluate \our from two aspects:
(1) long-horizon kinodynamic prediction accuracy and (2) closed-loop trajectory tracking navigation.
Experiments are conducted in the Verti-Bench simulator~\cite{xu2025verti} and the physical testbed similar to Verti-Arena~\cite{chen2025verti}. The simulation experiments assess generalization to new vehicle configurations and terrain conditions, while the physical experiments validate performance under real-world vehicle-terrain interactions. For closed-loop tracking, we specifically evaluate the benefit of online adaptation by comparing the adapted and fixed prior models under identical planning and control settings.

\subsection{Simulation Experiments in Verti-Bench}
\label{sec:sim_exp}

\subsubsection{Kinodynamic Prediction Accuracy}
\label{sec:sim_model}

We first evaluate whether the learned vehicle-terrain mobility descriptors provide an effective coefficient prior for new vehicle configurations. For each new vehicle, \our computes the SWD between its embedding distribution and the training mobility descriptors, selects the most relevant descriptors, and constructs a weighted coefficient prior. Prediction accuracy is evaluated over a $64$-step horizon. We compare \our against two baselines:

\begin{itemize}
    \item \textbf{From Scratch:} A vehicle-specific function encoder trained on the full set of $400$ trajectories collected from the new vehicle, serving as a data-intensive upper bound; and
    \item \textbf{Mobility Descriptors:} The coefficient set from each selected mobility descriptor is transferred directly to the new vehicle without any adaptation.
\end{itemize}

\begin{table}[h]
\centering
\caption{Kinodynamic prediction accuracy on novel vehicle configurations, denoted by $[\alpha_m,\mu_f,\alpha_s]$.}
\label{table:adapt}
\renewcommand{\arraystretch}{1.2}
\setlength{\tabcolsep}{7pt}
\small
\begin{tabular}{llcc}
\toprule
\textbf{Configuration} &
\textbf{Model} &
\textbf{SWD} &
\textbf{MSE $\pm$ Std} $\downarrow$ \\
\midrule

\multirow{5}{*}{[0.6, 0.75, 1.2]}
& \cellcolor[gray]{.9}\our 
& \cellcolor[gray]{.9}--
& \cellcolor[gray]{.9}0.144 $\pm$ 0.150 \\
& Descriptor 1 & 3.0406 & 0.201 $\pm$ 0.187 \\
& Descriptor 2 & 3.8738 & 0.238 $\pm$ 0.201 \\
& Descriptor 3 & 4.5326 & 0.224 $\pm$ 0.185 \\
& \textbf{From Scratch} & -- & \textbf{0.045 $\pm$ 0.140} \\
\midrule

\multirow{5}{*}{[1.6, 0.8, 1.6]}
& \cellcolor[gray]{.9}\our
& \cellcolor[gray]{.9}--
& \cellcolor[gray]{.9}0.055 $\pm$ 0.060 \\
& Descriptor 1 & 3.4404 & 0.084 $\pm$ 0.070 \\
& Descriptor 2 & 3.6182 & 0.092 $\pm$ 0.079 \\
& Descriptor 3 & 4.8106 & 0.103 $\pm$ 0.056 \\
& \textbf{From Scratch} & -- & \textbf{0.025 $\pm$ 0.084} \\
\midrule

\multirow{5}{*}{[0.6, 0.7, 0.8]}
& \cellcolor[gray]{.9}\our
& \cellcolor[gray]{.9}--
& \cellcolor[gray]{.9}0.098 $\pm$ 0.086 \\
& Descriptor 1 & 2.2614 & 0.100 $\pm$ 0.104 \\
& Descriptor 2 & 4.9236 & 0.104 $\pm$ 0.102 \\
& Descriptor 3 & 4.9886 & 0.110 $\pm$ 0.109 \\
& \textbf{From Scratch} & -- & \textbf{0.042 $\pm$ 0.118} \\
\bottomrule
\end{tabular}
% \vspace{-10pt}
\end{table}

As shown in Table~\ref{table:adapt}, \our outperforms direct transfer from mobility descriptors across all novel vehicle configurations, reducing prediction error by up to $34.52\%$. The results also show that descriptors with smaller SWD generally yield lower MSE, supporting SWD as an effective measure of transfer relevance. Although the ``From Scratch'' upper bound achieves the lowest error using 400 platform-specific trajectories, \our constructs an effective coefficient prior from only three trajectories. This demonstrates that the learned vehicle-terrain mobility representation enables data-efficient adaptation while preserving long-horizon prediction accuracy.

We further compare \our with MAML~\cite{finn2017model} and AnyCar~\cite{xiao2025anycar} to evaluate few-shot generalization. MAML performs gradient-based adaptation from a shared initialization, while AnyCar learns a universal dynamics model across multiple vehicle platforms. We evaluate both the original AnyCar model and a terrain-conditioned variant, denoted as AnyCar-Terrain. Each baseline is adapted using either three or $400$ trajectories from the novel configuration $[0.6,0.75,1.2]$, whereas \our uses only three trajectories to construct the coefficient prior.

\begin{table}[h]
\centering
\caption{Few-shot generalization comparisons over a $64$-step prediction horizon.}
\label{table:fewshot}
\renewcommand{\arraystretch}{1.3}
\setlength{\tabcolsep}{8pt}
\small
\begin{tabular}{lcc}
\toprule
\textbf{Model} & \textbf{New Vehicle Data} & \textbf{MSE $\pm$ Std} $\downarrow$ \\ 
\midrule

\rowcolor[gray]{.9}
\our
& 3 Trajectories
& \textbf{0.144 $\pm$ 0.150} \\
\midrule

\multirow{2}{*}{MAML}
& 3 Trajectories
& 0.330 $\pm$ 0.390 \\
& 400 Trajectories
& 0.319 $\pm$ 0.387 \\
\midrule

\multirow{2}{*}{AnyCar}
& 3 Trajectories
& 2.586 $\pm$ 1.548 \\
& 400 Trajectories
& 2.546 $\pm$ 1.536 \\
\midrule

\multirow{2}{*}{AnyCar-Terrain}
& 3 Trajectories
& 2.121 $\pm$ 1.257 \\
& 400 Trajectories
& 2.117 $\pm$ 1.254 \\
\bottomrule
\end{tabular}
% \vspace{-5pt}
\end{table}

Table~\ref{table:fewshot} shows that \our\ achieves the lowest long-horizon prediction error under the same few-shot setting. Using only three trajectories, it reduces MSE by $56.36\%$ relative to MAML, $94.43\%$ relative to AnyCar, and $93.21\%$ relative to AnyCar-Terrain. Although terrain conditioning improves AnyCar, both variants remain less accurate than \our. Moreover, increasing the adaptation data to $400$ trajectories provides limited improvement for MAML and AnyCar. These results demonstrate the data efficiency of transferring a structured coefficient prior from relevant vehicle-terrain mobility descriptors.

\subsubsection{Trajectory Tracking Navigation}
\label{sec:sim_nav}

We integrate \our into the MPPI planner and evaluate closed-loop navigation in a seen environment, used for function encoder data collection, and a previously unseen environment. We compare two variants of \our: one refines the coefficients online using RLS, while the other retains the fixed coefficient prior. Each setting is evaluated over five trials. We report success rate; traversal time; Hausdorff distance (HD) to the reference trajectory over successful trials only; and average roll and pitch across all trials. The reference trajectory specifies only the desired planar path in $x$ and $y$. HD measures planar tracking accuracy, and roll and pitch quantify vehicle stability during MPPI trajectory execution.

\begin{table}[h]
\caption{\textbf{Simulated Navigation Performance:} Success Rate, Traversal Time, Hausdorff Distance, Roll, and Pitch.}
\label{table:sim_nav}
\centering
\renewcommand{\arraystretch}{1.2}
\setlength{\tabcolsep}{7pt}
\small
\begin{tabular}{ccc}
\toprule[1pt]
\textbf{Seen Environment} &
\textbf{Adapted Prior} &
\textbf{Fixed Prior} \\
\midrule
Success Rate $\uparrow$
& \textbf{5/5}
& \textbf{5/5} \\
Traversal Time $\downarrow$
& \textbf{28.59s $\pm$ 0.29s}
& 28.62s $\pm$ 0.33s \\
HD $\downarrow$
& \textbf{0.94m $\pm$ 0.22m}
& 1.40m $\pm$ 0.88m \\
Roll $\downarrow$
& 4.35\textdegree $\pm$ 0.08\textdegree
& \textbf{4.21\textdegree $\pm$ 0.07\textdegree} \\
Pitch $\downarrow$
& \textbf{3.46\textdegree $\pm$ 0.04\textdegree}
& 3.51\textdegree $\pm$ 0.04\textdegree \\
\midrule
\textbf{Unseen Environment} &
\textbf{Adapted Prior} &
\textbf{Fixed Prior} \\
\midrule
Success Rate $\uparrow$
& \textbf{3/5}
& 0/5 \\
Traversal Time $\downarrow$
& \textbf{42.61s $\pm$ 5.31s}
& -- \\
HD $\downarrow$
& \textbf{4.72m $\pm$ 1.44m}
& -- \\
Roll $\downarrow$
& 5.50\textdegree $\pm$ 0.28\textdegree
& \textbf{4.62\textdegree $\pm$ 2.09\textdegree} \\
Pitch $\downarrow$
& \textbf{4.22\textdegree $\pm$ 1.11\textdegree}
& 6.51\textdegree $\pm$ 5.43\textdegree \\
\bottomrule[1pt]
\end{tabular}
% \vspace{-5pt}
\end{table}

As shown in Table~\ref{table:sim_nav}, both variants complete all trials in the seen environment, while the adapted prior reduces trajectory deviation from $1.40$\,m to $0.94$\,m with comparable traversal time and vehicle attitude stability. In the unseen environment, the adapted prior succeeds in three of five trials, whereas the fixed prior fails in all trials. These results demonstrate that online RLS refinement improves trajectory tracking and navigation robustness under unseen vehicle-terrain interactions.

\subsection{Physical Experiments in Testbed}
\label{sec:real_exp}

For real-world validation, we deploy \our on several open-source 1/10th-scale Verti-4-Wheeler platforms~\cite{datar2024toward}, with vehicle states measured at 100\,Hz via a motion-capture system. The training fleet comprises four configurations: Four-Wheeled Differential-Locked, Four-Wheeled Differential-Unlocked, Heavy-Payload, and Four-Tracked, with 20 minutes of trajectory data collected for each configuration. As the new platform, we introduce a Four-Wheeled with Trailer platform, which introduces coupled vehicle-trailer kinodynamics absent from the training fleet.

\subsubsection{Kinodynamic Prediction Accuracy}
\label{sec:real_model}

\begin{figure}[ht]
    \centering
    \includegraphics[width=\columnwidth]{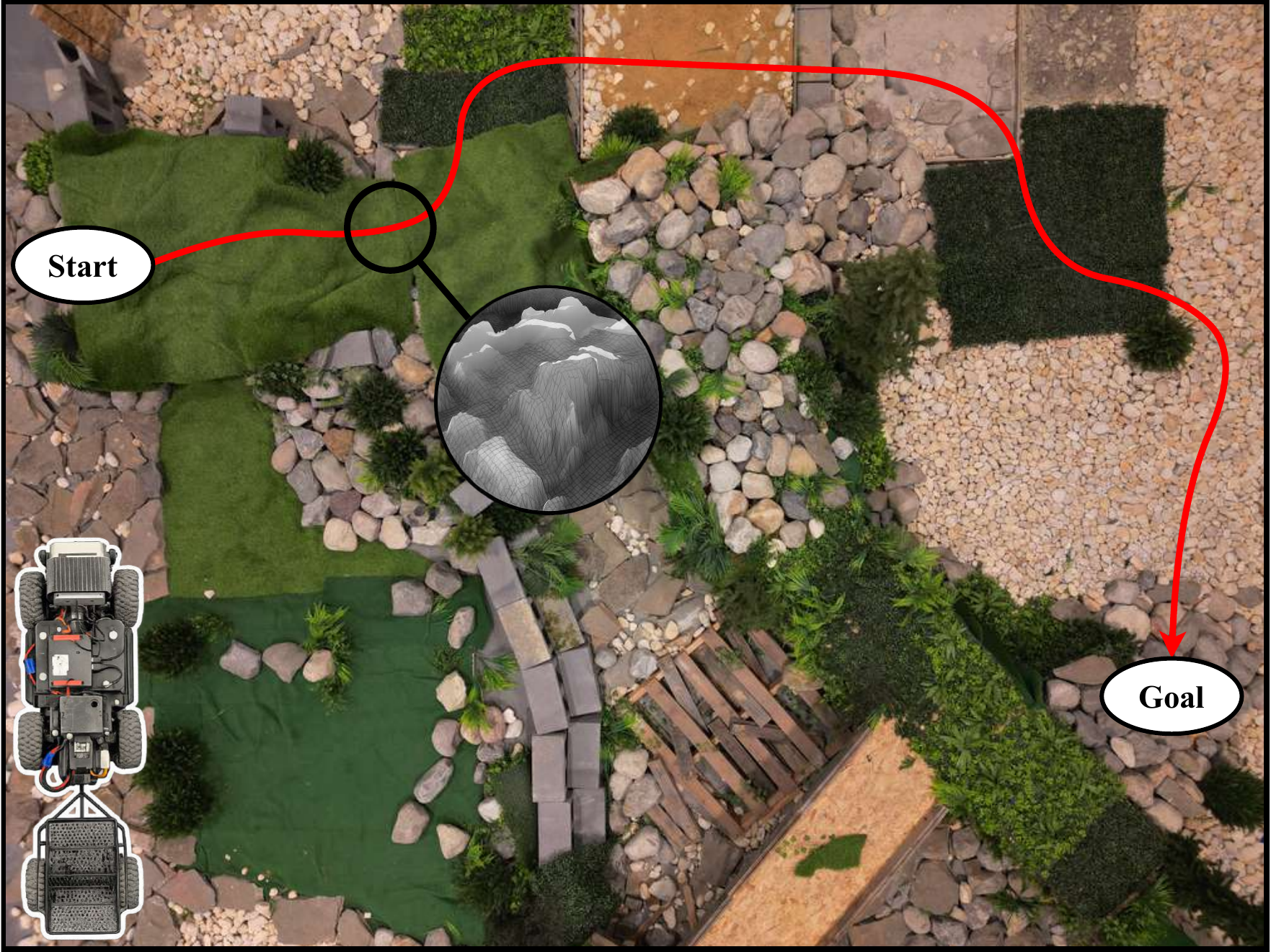}
    \caption{Physical testbed comprises a variety of off-road terrain, includes different geometries and semantics.}
    \label{fig::vert_arena}
    % \vspace{-10pt}
\end{figure}

We evaluate $64$-step prediction accuracy on the new physical configuration using trajectory and terrain data collected across diverse uneven terrain in physical testbed, as shown in Fig.~\ref{fig::vert_arena}.

\begin{table}[h]
\centering
\caption{Physical kinodynamic prediction accuracy on the novel configuration $[1.6,0.9,1.4]$.}
\label{table:real_model}
\renewcommand{\arraystretch}{1.2}
\setlength{\tabcolsep}{8pt}
\small
\begin{tabular}{lcc}
\toprule
\textbf{Model} &
\textbf{SWD} &
\textbf{MSE $\pm$ Std} $\downarrow$ \\
\midrule

\rowcolor[gray]{.9}
\our
& --
& 0.0077 $\pm$ 0.0051 \\
Descriptor 1
& 5.3315
& 0.0111 $\pm$ 0.0070 \\
Descriptor 2
& 6.2555
& 0.0160 $\pm$ 0.0104 \\
Descriptor 3
& 6.3464
& 0.0142 $\pm$ 0.0087 \\
Descriptor 4
& 7.6305
& 0.0201 $\pm$ 0.0123 \\
Descriptor 5
& 10.8642
& 0.0405 $\pm$ 0.0214 \\
\textbf{From Scratch}
& --
& \textbf{0.0051 $\pm$ 0.0036} \\
\bottomrule
\end{tabular}
\end{table}

Table~\ref{table:real_model} shows that \our reduces MSE by $30.63\%$ relative to the best mobility descriptor while approaching the performance of the vehicle-specific ``From Scratch'' model. This confirms that the learned physical vehicle-terrain descriptors provide an effective coefficient prior for downstream kinodynamic prediction on an unseen embodiment.

\subsubsection{Trajectory Tracking Navigation}
\label{sec:real_nav}

We finally evaluate closed-loop trajectory tracking in physical testbed over five trials, comparing the same two variants of \our and reporting the same performance metrics as in simulation.

\begin{table}[h!]
\centering
\caption{\textbf{Physical Navigation Performance:} Success rate, traversal time, roll, and pitch.}
\label{table:real_nav}
\renewcommand{\arraystretch}{1.2}
\setlength{\tabcolsep}{7pt}
\small
\begin{tabular}{ccc}
\toprule[1pt]
&
\textbf{Adapted Prior} &
\textbf{Fixed Prior} \\
\midrule
Success Rate $\uparrow$
& \textbf{5/5}
& 3/5 \\
Traversal Time $\downarrow$
& \textbf{42.68s $\pm$ 1.75s}
& 44.57s $\pm$ 4.08s \\
Roll $\downarrow$
& \textbf{4.92\textdegree $\pm$ 0.37\textdegree}
& 5.77\textdegree $\pm$ 0.88\textdegree \\
Pitch $\downarrow$
& 8.18\textdegree $\pm$ 0.40\textdegree
& \textbf{8.04\textdegree $\pm$ 1.24\textdegree} \\
\bottomrule[1pt]
\end{tabular}
% \vspace{-5pt}
\end{table}

Table~\ref{table:real_nav} shows that the adapted prior completes all five trials, compared with three successful trials using the fixed prior. Online RLS updates also reduce traversal time and roll, while maintaining comparable pitch. These results demonstrate that online coefficient refinement improves real-world navigation reliability.

\section{Conclusions and Limitations}

We present \our, a unified framework for cross-vehicle transfer and online adaptation of off-road kinodynamic knowledge over geometrically and semantically diverse terrain. By jointly encoding vehicle configurations, trajectory transitions, and terrain geometry and semantics, \our learns a structured vehicle-terrain mobility representation that supports relevant descriptor retrieval and weighted coefficient initialization. The resulting terrain-conditioned function encoder is refined online using RLS and integrated with MPPI for closed-loop navigation. With limited data from a new vehicle, \our reduces long-horizon prediction error by up to $34.52\%$ over direct mobility transfer and $94.43\%$ over competing baselines. Simulation and physical experiments further demonstrate reliable trajectory tracking under unseen vehicle-terrain interactions.

A key limitation of this work is that current kinodynamic model represents terrain effects through learned elevation and semantic features, making it less suitable for deformable terrain such as mud, sand, and snow, where vehicle motion depends on evolving wheel-terrain interactions and terrain deformation. To address this, future work will integrate terrain mechanic equations into a neuro-symbolic kinodynamic architecture, embedding physics-informed priors for sinkage, slip, and traction to enable robust online adaptation on deformable terrain.

\section*{Acknowledgments}
This work has taken place in the RobotiXX Laboratory at George Mason University. RobotiXX research is supported by National Science Foundation (NSF, 2350352), Army Research Office (ARO, W911NF2320004, W911NF2520011), Army Ground Vehicle Systems Center (GVSC), Google DeepMind (GDM), Microsoft Research (MSR), Clearpath Robotics, FrodoBots Lab, Raytheon Technologies (RTX), Tangenta, 4-VA, Mason Innovation Exchange (MIX), and Walmart.


%%%%%%%%%%%%%%%%%%%%%%%%%%%%%%%%%%%%%%%%%%%%%%%%%%%%%%%%%%%%%%%%%%%%%%%%%%%%%%%%

%%%%%%%%%%%%%%%%%%%%%%%%%%%%%%%%%%%%%%%%%%%%%%%%%%%%%%%%%%%%%%%%%%%%%%%%%%%%%%%%
\bibliographystyle{IEEEtran}
\bibliography{IEEEabrv,references}

\begin{thebibliography}{10}
\providecommand{\url}[1]{#1}
\csname url@samestyle\endcsname
\providecommand{\newblock}{\relax}
\providecommand{\bibinfo}[2]{#2}
\providecommand{\BIBentrySTDinterwordspacing}{\spaceskip=0pt\relax}
\providecommand{\BIBentryALTinterwordstretchfactor}{4}
\providecommand{\BIBentryALTinterwordspacing}{\spaceskip=\fontdimen2\font plus
\BIBentryALTinterwordstretchfactor\fontdimen3\font minus \fontdimen4\font\relax}
\providecommand{\BIBforeignlanguage}[2]{{%
\expandafter\ifx\csname l@#1\endcsname\relax
\typeout{** WARNING: IEEEtran.bst: No hyphenation pattern has been}%
\typeout{** loaded for the language `#1'. Using the pattern for}%
\typeout{** the default language instead.}%
\else
\language=\csname l@#1\endcsname
\fi
#2}}
\providecommand{\BIBdecl}{\relax}
\BIBdecl

\bibitem{xiao2022motion}
X.~Xiao, B.~Liu, G.~Warnell, and P.~Stone, ``Motion planning and control for mobile robot navigation using machine learning: a survey,'' \emph{Autonomous Robots}, vol.~46, no.~5, pp. 569--597, 2022.

\bibitem{wang2024survey}
N.~Wang, X.~Li, K.~Zhang, J.~Wang, and D.~Xie, ``A survey on path planning for autonomous ground vehicles in unstructured environments,'' \emph{Machines}, vol.~12, no.~1, p.~31, 2024.

\bibitem{borges2022survey}
P.~V. Borges, T.~Peynot, S.~Liang, B.~Arain, M.~Wildie, M.~G. Minareci, S.~Lichman, G.~Samvedi, I.~Sa, N.~Hudson \emph{et~al.}, ``A survey on terrain traversability analysis for autonomous ground vehicles: Methods, sensors, and challenges,'' \emph{Field Robotics}, vol.~2, pp. 1567--1627, 2022.

\bibitem{williams2018information}
G.~Williams, P.~Drews, B.~Goldfain, J.~M. Rehg, and E.~A. Theodorou, ``Information-theoretic model predictive control: Theory and applications to autonomous driving,'' \emph{IEEE Transactions on Robotics}, vol.~34, no.~6, pp. 1603--1622, 2018.

\bibitem{lavalle2001randomized}
S.~M. LaValle and J.~J. Kuffner~Jr, ``Randomized kinodynamic planning,'' \emph{The international journal of robotics research}, vol.~20, no.~5, pp. 378--400, 2001.

\bibitem{datar2024toward}
A.~Datar, C.~Pan, M.~Nazeri, and X.~Xiao, ``Toward wheeled mobility on vertically challenging terrain: Platforms, datasets, and algorithms,'' in \emph{2024 IEEE International Conference on Robotics and Automation (ICRA)}.\hskip 1em plus 0.5em minus 0.4em\relax IEEE, 2024, pp. 16\,322--16\,329.

\bibitem{xiao2021learning}
X.~Xiao, J.~Biswas, and P.~Stone, ``Learning inverse kinodynamics for accurate high-speed off-road navigation on unstructured terrain,'' \emph{IEEE Robotics and Automation Letters}, vol.~6, no.~3, pp. 6054--6060, 2021.

\bibitem{datar2024terrain}
A.~Datar, C.~Pan, M.~Nazeri, A.~Pokhrel, and X.~Xiao, ``Terrain-attentive learning for efficient 6-dof kinodynamic modeling on vertically challenging terrain,'' in \emph{2024 IEEE/RSJ International Conference on Intelligent Robots and Systems (IROS)}.\hskip 1em plus 0.5em minus 0.4em\relax IEEE, 2024, pp. 5438--5443.

\bibitem{cai2025pietra}
X.~Cai, J.~Queeney, T.~Xu, A.~Datar, C.~Pan, M.~Miller, A.~Flather, P.~R. Osteen, N.~Roy, X.~Xiao \emph{et~al.}, ``Pietra: Physics-informed evidential learning for traversing out-of-distribution terrain,'' \emph{IEEE Robotics and Automation Letters}, vol.~10, no.~3, pp. 2359--2366, 2025.

\bibitem{zhao2024physord}
Z.~Zhao, B.~Li, Y.~Du, T.~Fu, and C.~Wang, ``Physord: a neuro-symbolic approach for physics-infused motion prediction in off-road driving,'' in \emph{2024 IEEE/RSJ International Conference on Intelligent Robots and Systems (IROS)}.\hskip 1em plus 0.5em minus 0.4em\relax IEEE, 2024, pp. 11\,670--11\,677.

\bibitem{ward2025online}
W.~Ward, S.~Etter, T.~Ingebrand, C.~Ellis, A.~J. Thorpe, and U.~Topcu, ``Online adaptation of terrain-aware dynamics for planning in unstructured environments,'' \emph{arXiv preprint arXiv:2506.04484}, 2025.

\bibitem{ingebrand2024zero}
T.~Ingebrand, A.~J. Thorpe, and U.~Topcu, ``Zero-shot transfer of neural odes,'' \emph{Advances in Neural Information Processing Systems}, vol.~37, pp. 67\,604--67\,626, 2024.

\bibitem{xu2026vertiadaptor}
T.~Xu, C.~Pan, A.~Datar, and X.~Xiao, ``Vertiadaptor: Online kinodynamics adaptation for vertically challenging terrain,'' in \emph{2026 IEEE/RSJ International Conference on Intelligent Robots and Systems (IROS)}.\hskip 1em plus 0.5em minus 0.4em\relax IEEE, 2026.

\bibitem{chen2018neural}
R.~T. Chen, Y.~Rubanova, J.~Bettencourt, and D.~K. Duvenaud, ``Neural ordinary differential equations,'' \emph{Advances in neural information processing systems}, vol.~31, 2018.

\bibitem{Yang2024Pushing}
J.~Yang, C.~Glossop, A.~Bhorkar, D.~Shah, Q.~Vuong, C.~Finn, D.~Sadigh, and S.~Levine, ``Pushing the limits of cross-embodiment learning for manipulation and navigation,'' in \emph{Proceedings of Robotics: Science and Systems}, 2024.

\bibitem{doshi2025scaling}
R.~Doshi, H.~R. Walke, O.~Mees, S.~Dasari, and S.~Levine, ``Scaling cross-embodied learning: One policy for manipulation, navigation, locomotion and aviation,'' in \emph{Conference on Robot Learning}.\hskip 1em plus 0.5em minus 0.4em\relax PMLR, 2025, pp. 496--512.

\bibitem{xiao2025anycar}
W.~Xiao, H.~Xue, T.~Tao, D.~Kalaria, J.~M. Dolan, and G.~Shi, ``Anycar to anywhere: Learning universal dynamics model for agile and adaptive mobility,'' in \emph{2025 IEEE International Conference on Robotics and Automation (ICRA)}.\hskip 1em plus 0.5em minus 0.4em\relax IEEE, 2025, pp. 8819--8825.

\bibitem{xu2025verti}
T.~Xu, C.~Pan, M.~B. Rao, A.~Datar, A.~Pokhrel, Y.~Lu, and X.~Xiao, ``Verti-bench: A general and scalable off-road mobility benchmark for vertically challenging terrain,'' in \emph{Robotics: Science and Systems (RSS) 2025}, 2025.

\bibitem{wong2022theory}
J.~Y. Wong, \emph{Theory of ground vehicles}.\hskip 1em plus 0.5em minus 0.4em\relax John Wiley \& Sons, 2022.

\bibitem{atreya2022high}
P.~Atreya, H.~Karnan, K.~S. Sikand, X.~Xiao, S.~Rabiee, and J.~Biswas, ``High-speed accurate robot control using learned forward kinodynamics and non-linear least squares optimization,'' in \emph{2022 IEEE/RSJ International Conference on Intelligent Robots and Systems (IROS)}.\hskip 1em plus 0.5em minus 0.4em\relax IEEE, 2022, pp. 11\,789--11\,795.

\bibitem{datar2024learning}
A.~Datar, C.~Pan, and X.~Xiao, ``Learning to model and plan for wheeled mobility on vertically challenging terrain,'' \emph{IEEE Robotics and Automation Letters}, vol.~10, no.~2, pp. 1505--1512, 2024.

\bibitem{finn2017model}
C.~Finn, P.~Abbeel, and S.~Levine, ``Model-agnostic meta-learning for fast adaptation of deep networks,'' in \emph{International conference on machine learning}.\hskip 1em plus 0.5em minus 0.4em\relax PMLR, 2017, pp. 1126--1135.

\bibitem{xian2021hyperdynamics}
Z.~Xian, S.~Lal, H.-Y. Tung, E.~A. Platanios, and K.~Fragkiadaki, ``Hyperdynamics: Meta-learning object and agent dynamics with hypernetworks,'' \emph{arXiv preprint arXiv:2103.09439}, 2021.

\bibitem{kumar2021rma}
A.~Kumar, Z.~Fu, D.~Pathak, and J.~Malik, ``Rma: Rapid motor adaptation for legged robots,'' \emph{arXiv preprint arXiv:2107.04034}, 2021.

\bibitem{xu2026car}
T.~Xu, C.~Pan, and X.~Xiao, ``Car: Cross-vehicle kinodynamics adaptation via mobility representation,'' \emph{arXiv preprint arXiv:2603.06866}, 2026.

\bibitem{ortiz2024online}
K.~Ortiz, R.~DiPirro, A.~J. Thorpe, and M.~Oishi, ``Online learning of dynamical systems using low-rank updates to physics-informed kernel distribution embeddings,'' in \emph{2024 IEEE 63rd Conference on Decision and Control (CDC)}.\hskip 1em plus 0.5em minus 0.4em\relax IEEE, 2024, pp. 7548--7555.

\bibitem{wang2005gaussian}
J.~Wang, A.~Hertzmann, and D.~J. Fleet, ``Gaussian process dynamical models,'' \emph{Advances in neural information processing systems}, vol.~18, 2005.

\bibitem{peebles2023scalable}
W.~Peebles and S.~Xie, ``Scalable diffusion models with transformers,'' in \emph{Proceedings of the IEEE/CVF international conference on computer vision}, 2023, pp. 4195--4205.

\bibitem{kolouri2019generalized}
S.~Kolouri, K.~Nadjahi, U.~Simsekli, R.~Badeau, and G.~Rohde, ``Generalized sliced wasserstein distances,'' \emph{Advances in neural information processing systems}, vol.~32, 2019.

\bibitem{chen2025verti}
H.~Chen, A.~Datar, T.~Xu, F.~Cancelliere, H.~Rangwala, M.~B. Rao, D.~Song, D.~Eichinger, and X.~Xiao, ``Verti-arena: A controllable and standardized indoor testbed for multi-terrain off-road autonomy,'' 2026, pp. 133--138.

\end{thebibliography}

\end{document}